\documentclass[10pt,twocolumn,letterpaper]{article}

\usepackage[pagenumbers]{cvpr}      
\usepackage{graphicx}
\usepackage{amsmath}
\usepackage{amssymb}
\usepackage{booktabs}
\usepackage{textcomp}
\usepackage{multicol}
\usepackage{csquotes}
\usepackage{hyperref}
\usepackage{dcolumn}
\usepackage{arydshln}
\usepackage{enumitem}
\usepackage{balance}
\usepackage{combelow}
\usepackage{tabularx}
\usepackage{colortbl}
\usepackage{multirow}
\usepackage{makecell}
\usepackage[noend]{algpseudocode}
\usepackage{algorithmicx,eqparbox,array}

\usepackage{xcolor,colortbl}
\definecolor{Gray}{gray}{0.9}

\usepackage{arydshln}
\usepackage{xspace}
\makeatletter
\def\adl@drawiv#1#2#3{%
        \hskip.5\tabcolsep
        \xleaders#3{#2.5\@tempdimb #1{1}#2.5\@tempdimb}%
                #2\z@ plus1fil minus1fil\relax
        \hskip.5\tabcolsep}
\newcommand{\cdashlinelr}[1]{%
  \noalign{\vskip\aboverulesep
           \global\let\@dashdrawstore\adl@draw
           \global\let\adl@draw\adl@drawiv}
  \cdashline{#1}
  \noalign{\global\let\adl@draw\@dashdrawstore
           \vskip\belowrulesep}}
\makeatother
\newcommand{\methodname}{MEnsA\xspace}
\def\cvprPaperID{*****} 
\def\confName{CVPR}
\def\confYear{2023}
\usepackage[capitalize]{cleveref}
\crefname{section}{Sec.}{Secs.}
\Crefname{section}{Section}{Sections}
\Crefname{table}{Table}{Tables}
\crefname{table}{Tab.}{Tabs.}

\begin{document}

\title{\emph{\methodname}: Mix-up Ensemble Average for Unsupervised Multi Target Domain Adaptation on 3D Point Clouds}
\author{Ashish Sinha\\
Simon Fraser University\\
Burnaby, Canada\\
{\tt\small ashish\_sinha@sfu.ca}
\and
Jonghyun Choi\\
Yonsei University\\
Seoul, South Korea\\
{\tt\small jc@yonsei.ac.kr}
}


\maketitle
\begin{abstract}
   Unsupervised domain adaptation (UDA) addresses the problem of distribution shift between the unlabeled target domain and labelled source domain. 
    While the single target domain adaptation (STDA) is well studied in both 2D and 3D vision literature, multi-target domain adaptation (MTDA) is barely explored for 3D data despite its wide real-world applications such as autonomous driving systems for various geographical and climatic conditions. 
    We establish an MTDA baseline for 3D point cloud data by proposing to mix the feature representations from all domains together to achieve better domain adaptation performance by an ensemble average, which we call \emph{{\bf M}ixup {\bf Ens}emble {\bf A}verage} or {\bf \methodname}.
    With the mixed representation, we use a domain classifier to improve at distinguishing the feature representations of source domain from those of target domains in a shared latent space.
    In extensive empirical validations on the challenging PointDA-10 dataset, we showcase a clear benefit of our simple method over previous unsupervised STDA and MTDA methods by large margins (up to $17.10\%$ and $4.76\%$ on averaged over all domain shifts). 
    We make the code publicly available \href{https://github.com/sinAshish/MEnsA_mtda}{here}.
\end{abstract}

\section{Introduction}
\label{sec:intro}

For real-world applications ranging from a surveillance system to self-driving cars, deep learning (DL) for 3D data has made significant progress in a wide variety of tasks including classification, segmentation, and detection~\cite{feng2018gvcnn,guo2020pct,qi2017pointnet,you2019pvrnet,zhou2018voxelnet}.
Despite the impressive success of DL on 2D vision tasks, its success in 3D data regime involving point cloud data is yet limited by several factors as follows. 
First, as the point clouds usually do not come with color or textural information, it is not trivial to encode the visual appearances of the structure.
Second, annotation cost for 3D is more expensive than that in 2D; the annotation of 3D point clouds may require several rotations, which sometimes is non-trivial due to partial occlusions.
Third, the domain gap that arises from the difference in distribution between the original training data (source domain) and the deploying environment (target domain) is larger than that of 2D data owing to the characteristic of 3D geometry~\cite{huang2021generation}.

In this work, we address the challenge of reducing the domain gaps for 3D point cloud data, which alleviates the need for extensive annotation across all domains.
Specifically, we focus on unsupervised domain adaptation (UDA), that involves transferring knowledge from a label-rich domain, \ie, source domain to a label-scarce domain, \ie, target domain to reduce the discrepancy between source and target data distributions, typically by exploiting the domain-invariant features~\cite{kodirov2015unsupervised,gong2012geodesic,fernando2013unsupervised,yoo2016pixel}. 
Unfortunately, most of the existing literature on UDA primarily focuses on 2D data. 

\begin{figure*}
    \centering
    \subfloat[Single Target Domain Adaptation setup]{\includegraphics[width=0.4\textwidth]{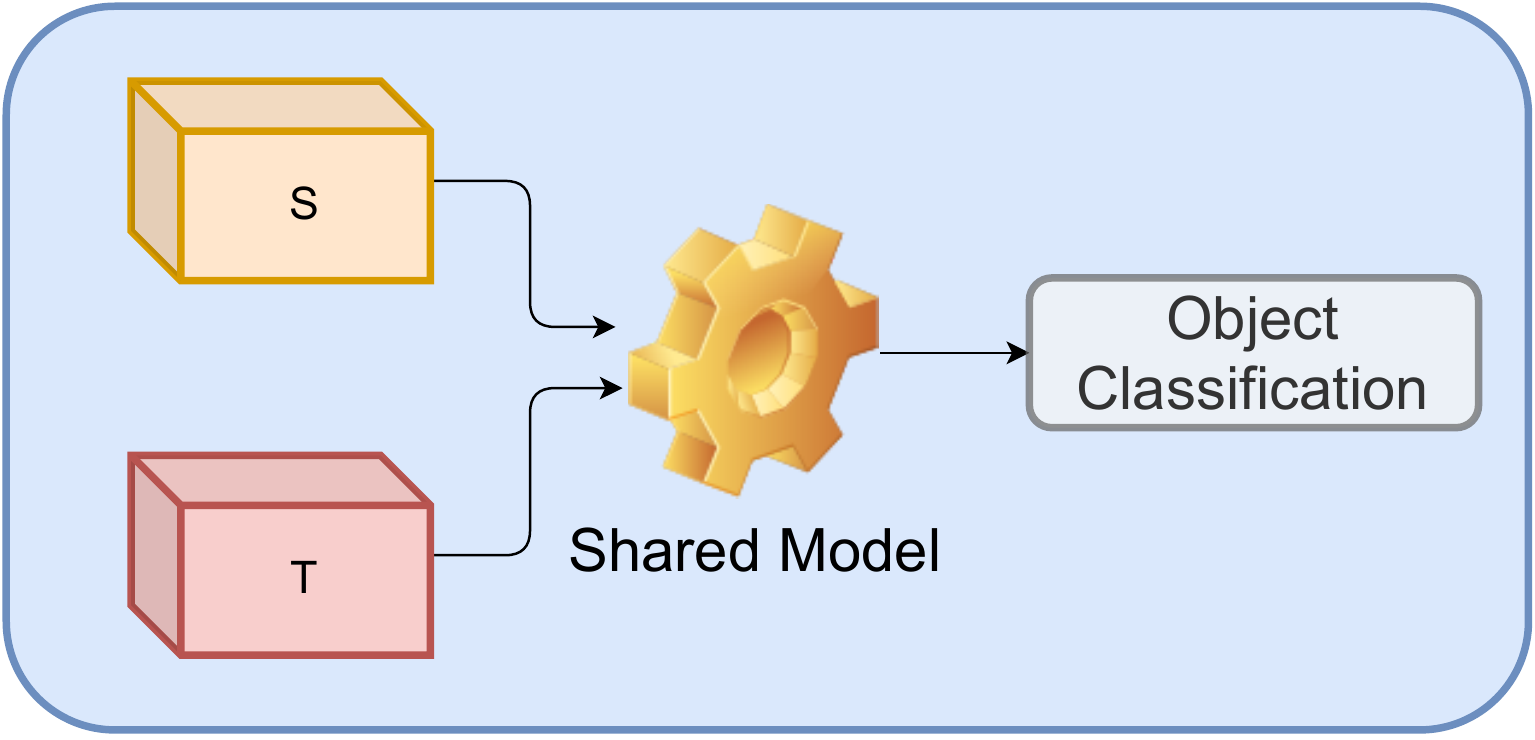}}
    \hspace{2em}
    \subfloat[Multi Target Domain Adaptation setup]{\includegraphics[width=0.38\textwidth]{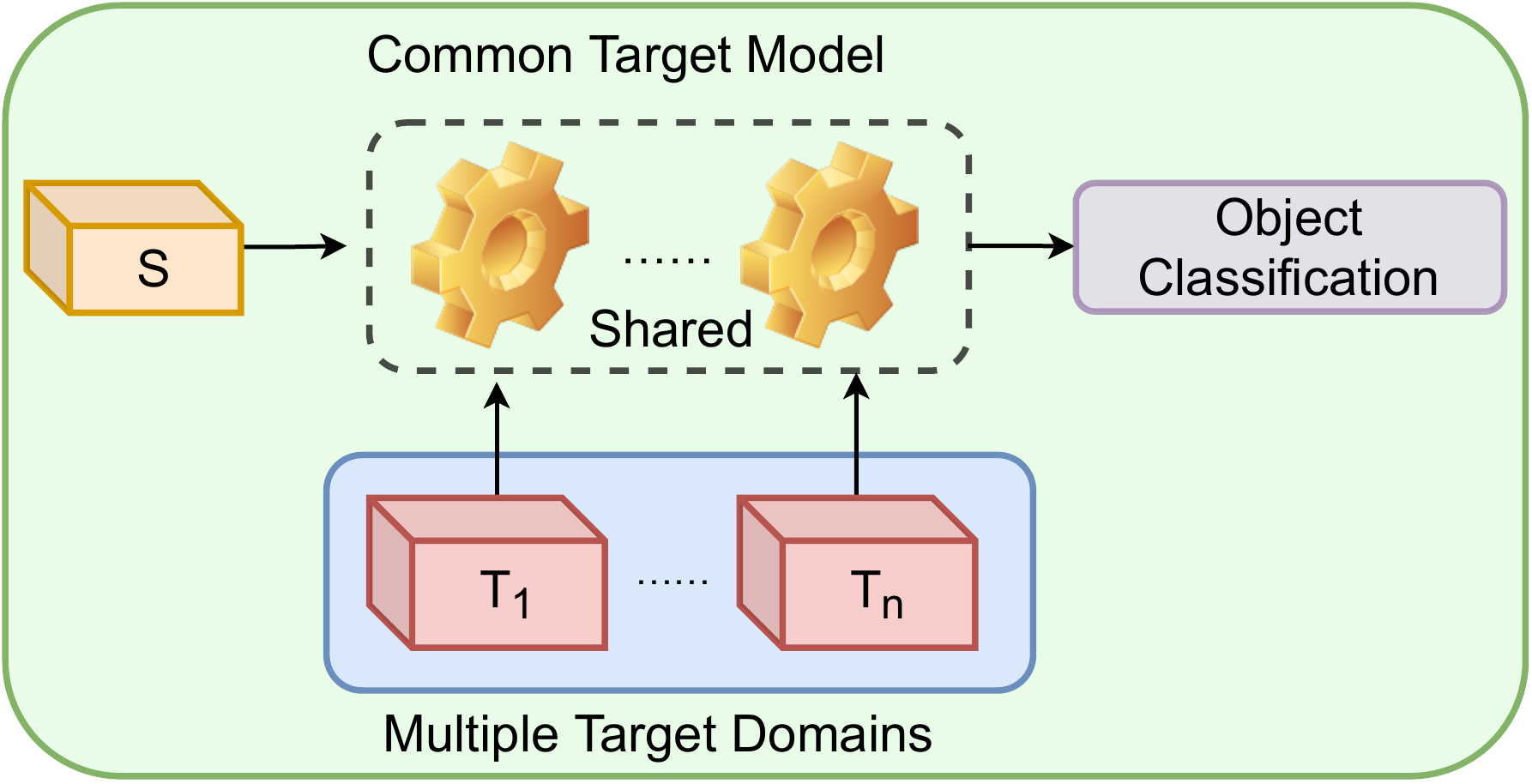}}
    \vspace{0.2em}
        \caption{Illustrative comparison of Single Target Domain Adaptation (STDA) and Multi Target Domain Adaptation (MTDA) setup. $S$ is the labelled source dataset, while $T_i$ are the unlabelled target datasets for $i = 1, 2,..., n$. STDA is a set-up where a single model is adapted to perform accurately on the target domain given a labelled source and unlabelled target data. MTDA is a set-up where a single model is adapted across all unlabelled target domains by learning on the labelled source domain. 
        }
    \label{fig:da_type}
\end{figure*}

The mortality risk and associated costs of conducting real-world experiments for autonomous driving and robotics systems have led to the increasing prevalence of synthetic data, particularly 3D data, in the research community~\cite{varol2017learning}. 
This necessitates the need to develop effective domain adaptation methods for 3D data across different domains, including real-to-sim or sim-to-real adaptation, to ensure successful deployment in real-world scenarios.

There are numerous works addressing the single-target domain adaptation (STDA) for 3D point clouds~\cite{qin2019pointdan,achituve2021self,huang2021generation}.
However, when 3D point cloud data of objects is collected under different environmental conditions using various depth cameras or LIDAR sensors for autonomous driving cars, it results in differences in statistical properties such as point cloud density, noise, and orientation. 
As a result, there is a pressing need for developing Multi-target Domain Adaptation (MTDA) methods, specifically for 3D point cloud data. 
Despite the well-studied 2D data regime~\cite{gholami2020unsupervised,nguyen2021unsupervised,chen2019blending}, MTDA in 3D point cloud domain remains an unexplored area in the literature.

In the context of both STDA and MTDA, if the category configurations are identical across all domains, one straightforward solution could be to extend STDA to MTDA by using one model per target domain. 
However, at inference time, it becomes challenging to determine the appropriate model to use when information about the target domain is not available. 
Moreover, as the number of target domains increases, the computational complexity increases accordingly.
Additionally, the model may experience catastrophic forgetting~\cite{serra2018overcoming, Wen2018OvercomingCF, Kirkpatrick2017OvercomingCF}, that involves a neural network trained on a particular task forgetting the previously learned information when trained on a new task.
As a result, the network's performance on the initial task deteriorates. 
This can be a significant challenge when adapting models to multiple target domains, as the model must be able to generalize well across all domains without forgetting the previously learned information.
Therefore, we argue that it is preferable to have \emph{a single model} that can adapt to multiple targets.
Hence, we propose to learn a single MTDA model for 3D point cloud. 
We illustrate the differences between STDA and MTDA in Figure~\ref{fig:da_type}.

To learn \emph{a single MTDA model}, we first model the multiple $N$ targets as a random variable. 
We then generate shared information between source and $N$ target domains as $N$ realizations of the shared representations by mixing them.
Then, we propose to take an ensemble average of the shared (\ie, mixed) representation for training a model that is invariant to multiple domains, calling it \emph{{\bf M}ixup {\bf Ens}emble {\bf A}verage} or {\bf \methodname}.
The shared representations are learned in a latent space for its low domain gaps~\cite{zhao2017multiple} in a min-max manner; maximizing the mutual information (MI) in the embedding space between the domains and domain-specific information while minimizing the MI between the domains and the domain-invariant information~\cite{gholami2020unsupervised}.
We show that our proposed method outperforms several STDA and MTDA approaches proposed for both 2D and 3D regimes on the multiple target domains evaluated on challenging PointDA-10 benchmark dataset~\cite{qin2019pointdan} by large margins.
In summary, we present the following contributions:
\begin{itemize}
    \item We show that a straightforward extension of domain adaptation methods designed for STDA, in particular 2D data, is non-trivial and does not transfer well to MTDA, specifically in the case of 3D data. 
    \item We propose a simple and novel ensemble-average based mixup approach, named \methodname, to address the challenging yet unaddressed task of adapting a \emph{single} model across multiple target domains by learning on a single source domain, on point cloud data.
    \item Extensive validations on PointDA-10 dataset demonstrates a significant benefit of our simple approach over previous unsupervised STDA and MTDA methods by large margins (up to 17.10\% and 4.76\% on averaged over all domain shifts).
    \item To the best of our knowledge, this is the first work that benchmarks and addresses the task of MTDA on 3D data, specifically 3D point clouds.

\end{itemize}

\section{Related Work}
\label{sec:rel_work}

\subsection{3D Point Clouds}

3D visual data is represented in various ways; 3D mesh, voxel grid, implicit surfaces and point clouds. 
Deep neural networks (DNNs) have been employed to encode the different modalities of 3D data~\cite{su2015multi,you2018pvnet,feng2019meshnet,maturana2015voxnet,michalkiewicz2019implicit}. 
Among them, point clouds, represented by a set of $\{x, y, z\}$ coordinates, is the most straightforward modality to represent 3D spatial information. 
PointNet \cite{qi2017pointnet} was the pioneering model to encode point clouds, taking advantage of a symmetric function to obtain the invariance of point permutation. 
But it ignores the local geometric information, which may be vital for describing the objects in 3D space. 
PointNet++ \cite{qi2017pointnet++} proposed to stack PointNets hierarchically to model neighborhood information and increase model capacity. 
PointCNN \cite{li2018pointcnn} proposed $\mathcal{X}$-Conv to aggregate features in local patches and apply a bottom-up network structure like typical CNNs. 
Recent works~\cite{guo2020pct,zhao2020point} propose to attend to point-point interactions using self-attention layers and achieve state-of-the-art accuracy on ``supervised" classification and segmentation tasks. 

Despite the wide usage, point cloud data suffers from labelling efficiency. 
In real-world scenario, some parts of an object may be occluded or lost (\eg, chairs lose legs) while scanning from acquisition devices, \eg, LIDAR, making annotation difficult. 
To alleviate the annotation cost, unsupervised domain adaption (UDA) method for point clouds could be a remedy. 

\subsection{Single Target Domain Adaptation (STDA)}

STDA is an unsupervised transfer learning approach which focuses on adapting a model to perform accurately on unlabeled target data while using labelled source data. 
Most of the prior works are proposed for 2D data~\cite{long2013transfer, ganin2015unsupervised, tzeng2017adversarial, sun2017correlation}. 
They are categorized as (1) adversarial, (2) discrepancy, and (3) reconstruction-based approaches. 
The adversarial approach refers to a model with a discriminator and a generator, where the generator aims to fool the discriminator until the discriminator is unable to distinguish the generated features between the two domains \cite{ganin2015unsupervised,tzeng2017adversarial,qin2019generatively,dong2019semantic}. 
These approaches have been proposed using either gradient reversal \cite{ganin2015unsupervised} or a combination of feature extractor and domain classifier to encourage domain confusion. 
The discrepancy based approaches \cite{long2013transfer} rely on measures between source and target distributions that can be minimized to generalize on the target domain. 
The reconstruction-based approaches focus on the mapping of the source domain to the target domain data or vice versa \cite{bousmalis2017unsupervised,hong2018conditional}. 
They often rely on the use of GAN \cite{goodfellow2014generative} in order to find a mapping between source and target.

The STDA methods for 3D point clouds include a self-adaptive module for aligning local features \cite{qin2019pointdan}, deformation reconstruction as a pretext task \cite{achituve2021self} or generating synthetic data from source domain to closely match data from target domain \cite{huang2021generation}. 
Recent works \cite{zhou2018unsupervised,xu2020adversarial,huang2021generation,saleh2019domain,achituve2021self} have been proposed which either use an augmentation method as a self-supervised task or generate synthetic data from source domain to mimic the target domain for UDA on point-clouds in a STDA setting.
Nevertheless, extending these approaches in MTDA scenario is not straightforward.

\subsection{Multiple Target Domain Adaptation (MTDA)}

MTDA requires adapting a model to perform accurately across multiple unlabeled target domains using labelled data from a single source domain. 
However, the existing MTDA literature has primarily focused on 2D data~\cite{chen2019blending, nguyen2021unsupervised, gholami2020unsupervised}, where they either use target domain labels \cite{gholami2020unsupervised} or not~\cite{chen2019blending,liu2020open,peng2019domain,nguyen2021unsupervised}. 
Gholami \etal~\cite{gholami2020unsupervised} proposed an approach to adapt to multiple target domains by maximizing the mutual information between domain labels and domain-specific features while minimizing the mutual information between the shared features. 
Chen \etal~\cite{chen2019blending} proposed to blend multiple target domains together and minimize the discrepancy between the source and the blended targets. 
Liu \etal~\cite{liu2020open} proposed to use a curriculum learning based domain adaptation strategy combined with an augmentation of feature representation from a source domain to handle multiple target domains. 
Nguyen \etal~\cite{nguyen2021unsupervised} proposed to perform UDA by exploiting the feature representation learned from different target domains using multiple teacher models and then transferring the knowledge to a common student model to generalize over all target domains using knowledge distillation. 
Although effective on 2D vision tasks, these methods often fail to generalize well on 3D vision tasks due to their design that focuses on images, and disregards local geometric information, and the problem of catastrophic forgetting that can occur during alternate optimization
~\cite{nguyen2021unsupervised}.
Consequently, MTDA for 3D vision tasks remains an underexplored research area despite its numerous real-world applications. 
Thus, we propose the first MTDA method for 3D point cloud.

\section{Approach}
\label{sec:appr}

\subsection{Overview}

Ganin \etal~\cite{ganin2015unsupervised} argues that representations that are indistinguishable between the source and target domains are crucial for domain invariant inference.
In the context of image classification~\cite{yun2019cutmix,zhang2017mixup}, a common data augmentation technique known as ``mixing" or linear interpolation of two images has been employed to make two samples indistinguishable from each other.
However, when considering domain-invariance of point clouds, directly mixing the input point clouds presents a challenge, as not all points are \emph{equally} important in describing the object, and it is not trivial to determine which points to mix and which points to exclude. 
Instead, we encode the point clouds using a DNN, which implicitly weighs the important points and their point-point interactions, and use the embeddings for mixing.
As argued in~\cite{yun2019cutmix, zhang2017mixup}, mixing can act as an effective regularizer for guiding the model to be discriminant of source domain from the target domains for point clouds, while remaining indiscriminant of the domain shifts across multiple domains.
This enables a model to generalize across multiple domains. 

\begin{figure*}[t]
    \centering
    \includegraphics[width=0.80\textwidth]{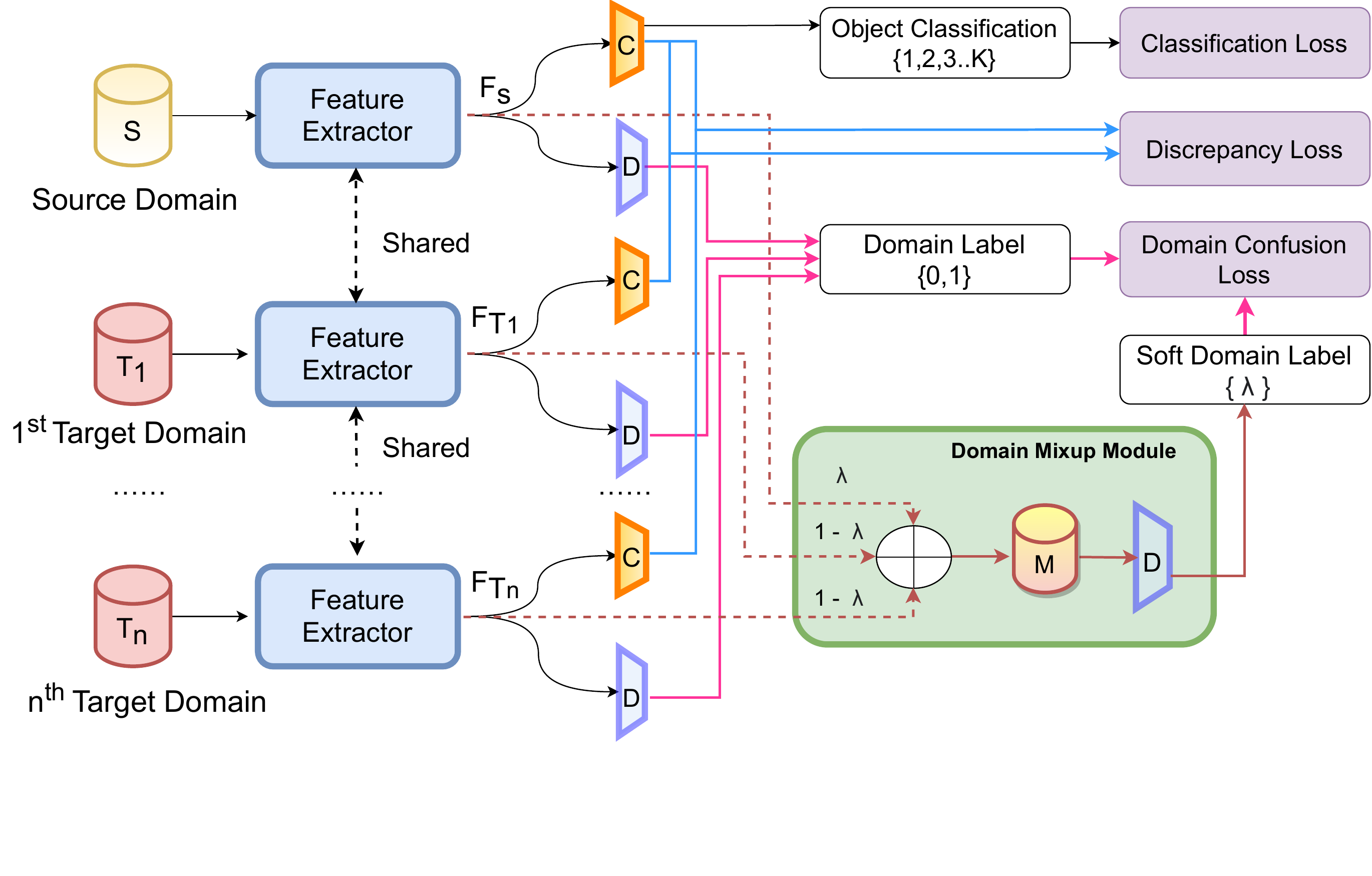}
    \vspace{-3em}
    \caption{\textbf{Overview of our MTDA model.} The labeled source data $S$, and the unlabeled target data $T_i$ from multiple domains $i=1..n$, are taken as input by the feature extractor. The source feature $F_s$ is used by object classifier $C$ and domain classifier $D$ to predict the category label, and domain of the input resp. The target feature $F_{T_i}$ is used by $C$ to calculate the discrepancy loss between the source and target features. $F_{T_i}$ is also used by $D$ to differentiate between source and target domain. $F_s$ and $F_{T_i}$ are fed to the {domain mixup module} to get mixed domain features $M$ to predict the soft scores for source/ target. The model is optimized using a combination of object classification loss, {domain confusion loss} and {discrepancy loss}.
    }
    \label{fig:arc}
\end{figure*}



We illustrate the overview of our proposed MTDA approach \methodname in Figure \ref{fig:arc}.
We employ an adversarial training strategy~\cite{ganin2015unsupervised} to reduce the distribution shifts across multiple domains, using gradient reversal for the \emph{domain confusion loss}. 
Specifically, we first encode the point clouds by the feature extractor module $F$ using a variant of the node attention module proposed in PointDAN~\cite{qin2019pointdan}.
This module $F$ preserves both local geometric structures and the global relations between the local features, resulting in a tensor $F_T$ that is split into two branches. 
The first branch, a \emph{domain classifier} $D$, is composed of a Gradient Reversal Layer (GRL) \cite{ganin2015unsupervised} and a fully connected layer. 
The GRL helps in building a feature representation of the raw input $\mathcal{X}$ that is good enough to predict the correct object label $\mathcal{Y}$, but such that the domain label of $\mathcal{X}$ cannot be easily deduced from the feature representation. 
This promotes domain confusion, where the feature extractor $F$ attempts to confuse the domain classifier $D$ by bridging the two distributions closer. 
The second branch is an object classifier $C$ consisting of a fully connected layer and a SoftMax activation function. 
$D$ uses $F_T$ to classify the feature representations into source or target domain, while $C$ classifies them into $K$ classes. 
Thus, $F$ is adversarially trained by minimizing the object classifier's classification and maximizing the domain classifier's classification loss.
Our model's core is the \emph{domain mix-up module}, which is explained in detail in the following section.

\subsection{Domain Mixup Module}
\label{sec:mixup}

Inspired by the mixup \cite{zhang2017mixup} approach for 2D data, we propose to mix the feature embeddings obtained by $F$, \emph{but} from multiple domains in the latent space.
Unlike the methods for 2D data where the input images are blended by an alpha factor \cite{yun2019cutmix,chen2020pointmixup}, we propose mixing the feature embeddings, since the feature embeddings from the deeper layers of the network contains information about the global shape of the point cloud and local point-point interaction, as demonstrated in~\cite{xu2020adversarial} applied to STDA set-up. 
Specifically, we linearly interpolate the source ($F_s$) and target feature ($F_{T_i}$) embeddings to obtain $F_i^m$ and the corresponding mixed \emph{soft} domain labels $L_i^m$ as:
\begin{equation}
    F_i^m = \lambda F_s + (1-\lambda) F_{T_i},
    \label{eq:main_f}
\end{equation}
\begin{equation}
    L_i^m = \lambda L_s + (1-\lambda) L_{T_i},
    \label{eq:main_l}
\end{equation}
where $L_s$ and $L_{T_i}$ denote the domain labels of source and target domain which are set to $1$ and $0$, respectively. 
The use of soft labels is essential in creating a continuous probability distribution that indicates the likelihood of a sample belonging to a particular domain.
Unlike hard domain labels that limit the classification of samples to just one domain, soft labels promote the learning of domain-invariant features that are useful for both domains and not biased towards one or the other.

The linear interpolation of feature embeddings serves two purposes.
Firstly, it helps create a continuous domain-invariant latent space, enabling the mixed features to be mapped to a location in-between the latent space of source and target domain \cite{berthelot2018understanding}.
This continuous latent space is crucial for domain-invariant inference across multiple domains.
Secondly, it acts as an effective regularizer, helping the domain classifier $D$ improve in predicting the soft scores for domains (source or target) of the mixed feature embeddings $F_i^m$,
similar to~\cite{yun2019cutmix, zhang2017mixup}.
Since our approach involves multiple target domains, we model domain invariant representation obtained by the mixup $F_i^m$ as a random variable.
By using multiple realizations of the `mixup' representation for different domains, we learn domain-invariant information that is robust to domain shifts.

\begin{figure*}
    \subfloat[Baseline Mixup Method (\textbf{Sep.})]{\includegraphics[width=0.475\textwidth]{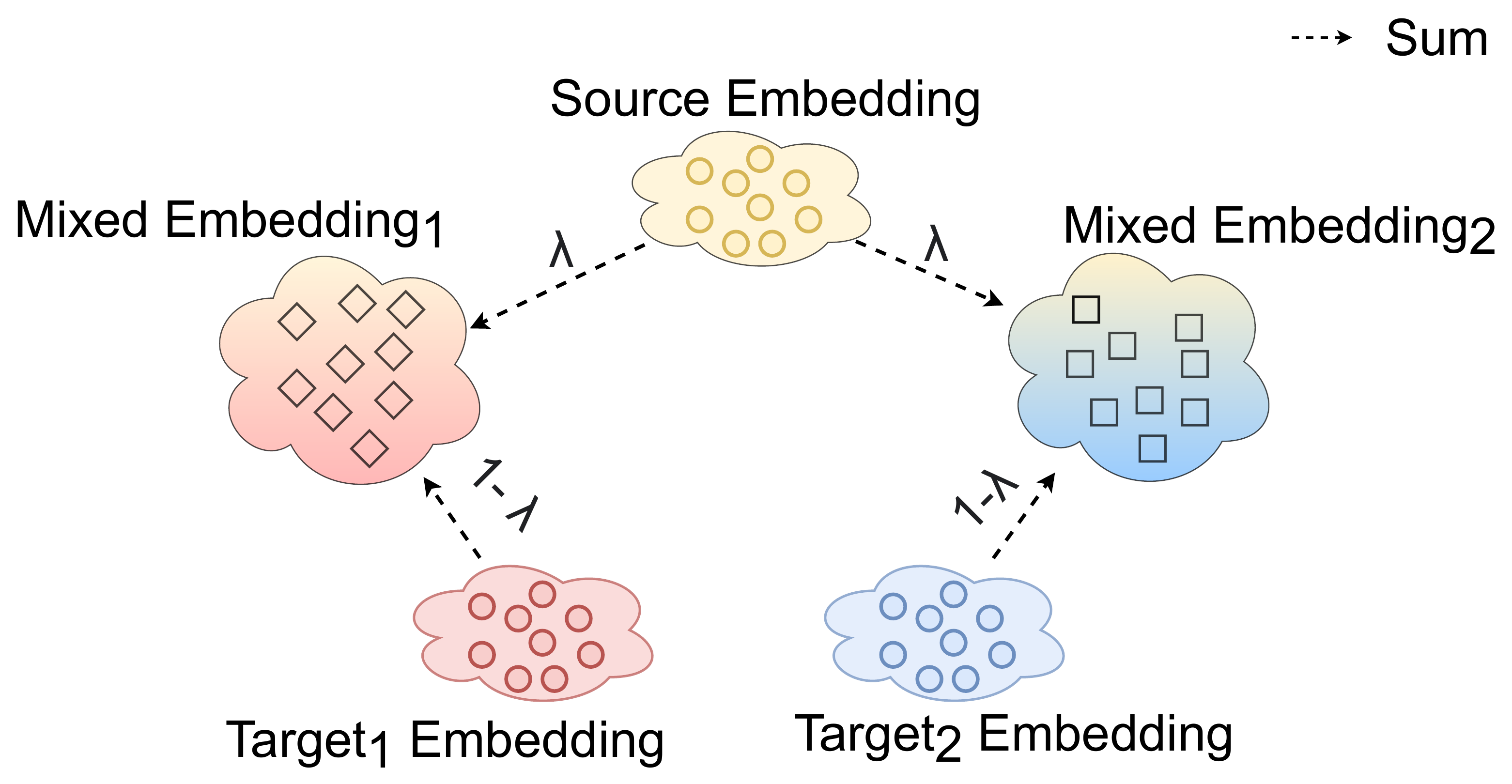}}
    \subfloat[Proposed (\textbf{MEnsA}) Method]{\includegraphics[width=0.475\textwidth]{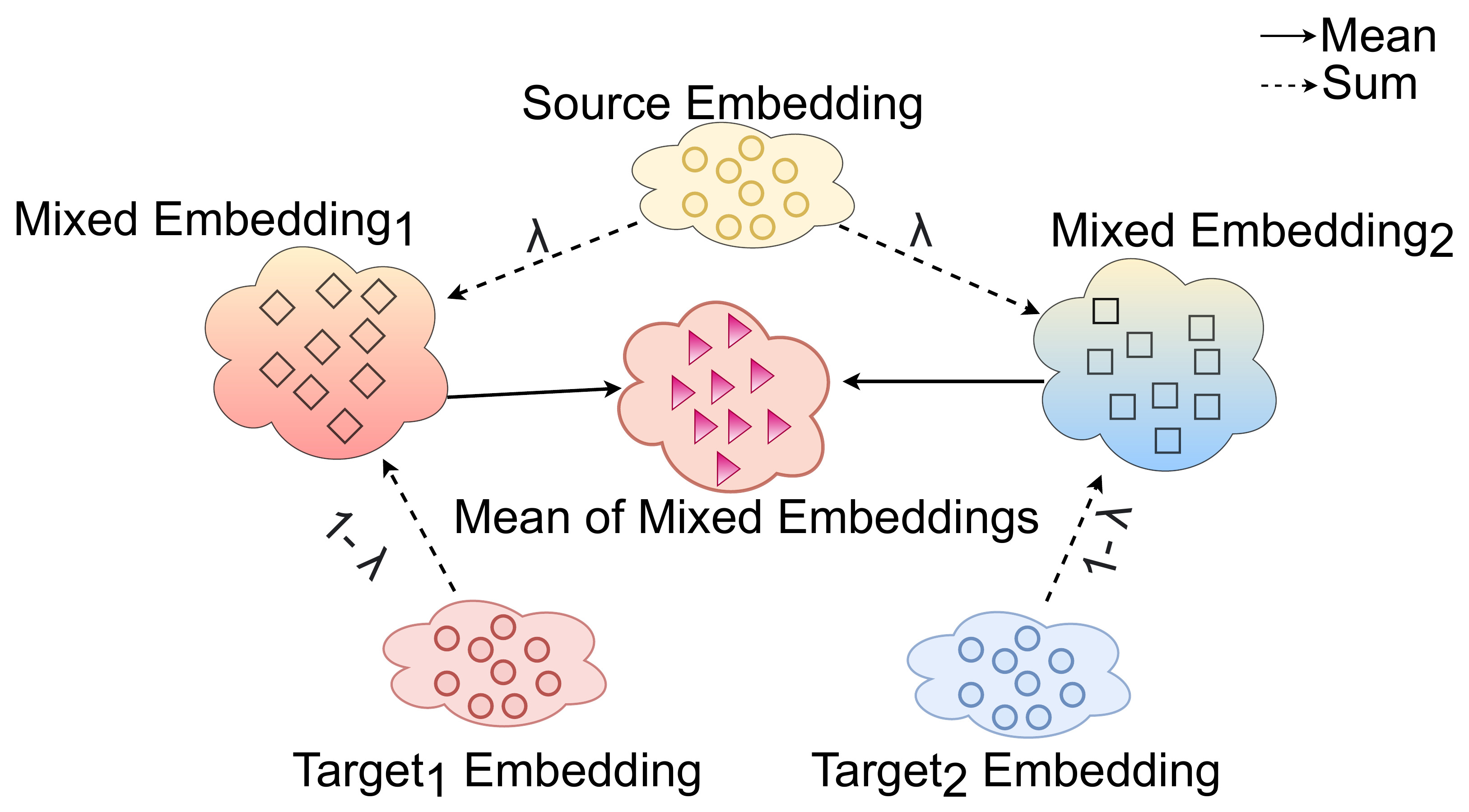}}
    \caption{\textbf{Comparative illustration of the mixup methods of proposed \textbf{`\methodname'} to a baseline method (\textbf{`Sep'})}. They mix feature embeddings of source and $N$ target domains (here, we use $N=2$ for visualization clarity). 
    The \textbf{Sep.} mixup method mixes source domain embeddings to each of the target domain embeddings to create $n$ mixed embeddings, $F_i^m$. 
    Each of which is passed to domain classifier $D$ to predict soft scores (mixup ratio between source and target domain) instead of hard labels for the domains. 
    }
    \label{fig:mixup}
\end{figure*}



\paragraph{Baseline mixup (Sep).} 
The standard approach for utilizing the stochastic realization of mixed embeddings, involves mixing the feature embeddings of the source domain $S$ and each of the target domains $T_i$ from a set of target domains $\mathcal{T}$ to train a model.
Specifically, each mixup feature is fed into the domain classifier $D$ separately for each of the target domains $\mathcal{T}$, which predicts a soft score, \ie, the mixup ratio for source $S$ and target domain $T_i$.
Then, the cross-entropy loss is calculated and back-propagated over the Gradient Reversal Layer (GRL).
We call this approach as the `Sep.' method and is illustrated in Figure
\ref{fig:mixup} 
(a).

\paragraph{Mixup Ensemble Average (\methodname).}
The sequential training approach employed in the Sep. method may not allow the model to effectively learn the interaction between the source and multiple target domains due to catastrophic forgetting~\cite{mccloskeyC89, serra2018overcoming}, as the method performs a pair-wise mixup between the source and target domains.
This results in the model forgetting previously learned domain-invariant features when exposed to a new target domain.
%
To alleviate this problem, we propose a simple method of taking an \emph{ensemble average} of the mixed feature embeddings from the multiple targets $F_i^m$ as:
\begin{equation}\label{eq:mensa}
    F_m^{M} = \frac{1}{n}\sum_{i=1}^{n} F_i^m.
\end{equation}
We call it \emph{{\bf M}ixup {\bf Ens}emble {\bf A}verage} or {\bf \methodname}, illustrated in Figure 
\ref{fig:mixup} (b).
The soft scores for the source and target domains are obtained by feeding the mixed feature $F_i^m$ to the domain classifier $D$, and the mapping between the source and each target domain is optimized by reproducing kernel Hilbert space (RKHS) \ie MMD.
We posit that the ensemble average effectively captures shared information across \emph{all} domains while mitigating conflicting information among them.
Consequently, the model trained on this averaged representation, captures differences between the source domain and multiple target domains in a consolidated manner, resulting in improved generalization over domain shifts across multiple target domains.

Our method differs from \cite{xu2020adversarial} in that they propose a pairwise mixup at the input and intermediate stage followed by the reconstruction of image samples. 
In contrast, we explore mixing in a 3D MTDA setup by mixing the latent features from all domains into one, rather than pairwise mixing. 
Our approach is designed to capture shared domain-invariant features across multiple domains, whereas pairwise mixup only focuses on learning domain-invariant features between the source and one target domain, ignoring the shared features across multiple domains, thereby suffering from catastrophic forgetting.

\subsection{Objective Function}
\label{sec:loss}
The complete architecture is trained end-to-end by minimizing $\mathcal{L}$, which is a weighted combination of supervised classification loss on the source domain ($\mathcal{L}_{cls}$), domain confusion loss ($\mathcal{L}_{dc}$), mixup loss ($\mathcal{L}_{mixup}$) and MMD loss  ($\mathcal{L}_{mmd}$), defined as:
\begin{equation}
    \mathcal{L} =\log \left( \sum ( e^ {\gamma (\mathcal{L}_{cls} + \eta \mathcal{L}_{dc} + \zeta \mathcal{L}_{adv})} ) \right) / \gamma,
    \label{eq:loss}
\end{equation}
Here, $\eta$, $\zeta$ and $\gamma$ are balancing hyperparameters. 
The classification, domain confusion and adversarial loss are cross-entropy losses, defined as:
\begin{equation}
\begin{split}
    \mathcal{L}_{cls} &= \mathcal{L}_{CE}(C(F_s), y_s), \\
    \mathcal{L}_{dc}  &= \mathcal{L}_{CE} (D(F_s), L_s) + \mathcal{L}_{CE} (D(F_{T_i}, L_{T_i}) ),\\
    \mathcal{L}_{adv} &= \lambda_{1} \mathcal{L}_{mmd} + \lambda_2 \mathcal{L}_{dc} + \lambda_3 \mathcal{L}_{mixup},
\end{split}
\label{eq:l_subs}
\end{equation}

where $C$ is the object classifier, $D$ is the domain classifier, $y_s$ is the ground truth object label, $L_s$ is the domain label for source and $L_{T_i}$ is the target domain label set as 1 and 0.
$\lambda_1, \lambda_2$ and $\lambda_3$ are balancing hyperparameters with constant values of $5.0$, $5.0$ and $1.2$ respectively, and are chosen empirically.

The MMD loss and mixup loss are defined as:
\begin{equation}
\begin{split}
    \mathcal{L}_{mmd} = \mathcal{L}_{rbf} ( C(F_s), F_{T_i}, \sigma),\\
    \mathcal{L}_{mixup} = \mathcal{L}_{CE} (D(F_m^{M}), L_i^m),
\end{split}
\end{equation}
where $\mathcal{L}_{rbf}$ is a radial basis function.


\section{Experiments}
\label{sec:exp}

\subsection{Experimental Set-up}
\label{sec:exp_setup}

\paragraph{Dataset.}
We evaluate our method on PointDA-10, a benchmark dataset proposed by \cite{qin2019pointdan} for the task of point cloud classification.
PointDA-10 consists of three subsets of three widely used datasets: ShapeNet \cite{chang2015shapenet}, ScanNet \cite{dai2017scannet} and ModelNet \cite{wu20153d}, each containing $10$ common classes (chair, table, monitor, etc.). 
\textbf{ModelNet-10 (M)}, called ModelNet hereafter, contains 
samples of clean 3D CAD models.
\textbf{ShapeNet-10 (S)}, called ShapeNet hereafter, contains 
samples of 3D CAD models collected from online repositories. 
\textbf{ScanNet-10 (S*)}, called ScanNet hereafter, contains 
samples of scanned and reconstructed real-world indoor scenes.
Samples from this dataset are significantly harder to classify because (1) many objects have missing parts due to occlusion, and (2) some objects are sampled sparsely.
For more details, we refer the readers to the supplementary material.

\paragraph{Implementation Details.}
\label{sec:implementation}

The proposed approach is implemented on PyTorch \cite{paszke2019pytorch} framework with Adam \cite{kingma2014adam} as the optimizer for training. The learning rate is assigned as $10^{-3}$ under the weight decay of $5^{-4}$ with $\beta_1$ and $\beta_2$ kept as $0.9$ and $0.999$. All models were trained for $100$ epochs with a batch size of $64$.
We set $\lambda_1, \lambda_2$ and $\lambda_3$ used in Equation~\ref{eq:l_subs} to $5.0$, $5.0$ and $1.2$ respectively. 
For Equation \ref{eq:main_f} and \ref{eq:main_l}, $\lambda \in  [0, 1]$ is a mixup ratio and $\lambda \sim \beta(\alpha, \alpha)$, where $\beta$ is a beta function and $\alpha$ is set to $2.0$ for all experiments. 
We sample $\lambda$ from a beta distribution, $\beta(\alpha_1, \alpha_2)$ such that $\alpha_1 $ = $\alpha_2$, as it enables sampling values from a non-skewed distribution. 

Motivated by \cite{nguyen2021unsupervised}, we use scheduled tuning for $\eta$ in Equation~\ref{eq:loss} as:
\begin{equation}\label{15}
    \eta = s \cdot e^ { \left( \frac{\log \frac{f}{s}}{N_e} \cdot e \right)}.
\end{equation}
where $s$ is the starting value of $0.1$, $f$ is the final value of $0.9$, $N_e$ is the total number of epochs and $e$ is the current epoch.
This helps in measuring the importance of domain confusion loss over time to adversarially raise the error rate of the domain classifier, thereby forcing it to improve at distinguishing the domains over time.


\paragraph{Baselines.}
\label{sec:baseline}
We compare the proposed approach with general purpose UDA methods including maximum mean discrepancy (MMD) \cite{long2013transfer}, adversarial discriminative domain adaptation (ADDA) \cite{ganin2015unsupervised}, domain adversarial neural network (DANN) \cite{tzeng2017adversarial} and maximum classifier discrepancy (MCD) \cite{saito2018maximum}. 
It is also compared with STDA method on point clouds \cite{qin2019pointdan}. 

We also compared our approach to MTDA approaches for 2D vision tasks involving blending targets \cite{chen2019blending}, exploiting shared and private domain spaces \cite{gholami2020unsupervised} and knowledge distillation from multiple teachers to a common student model \cite{nguyen2021unsupervised} with minor modification to use 3D point cloud data. 
In adapting these methods in MTDA scenario, we follow the authors' implementations and the hyperparameters are kept the same as proposed in the respective papers. 
Since \cite{nguyen2021unsupervised} was proposed for MTDA on 2D vision, the authors used ResNet50 \cite{he2016deep} as the teacher model and AlexNet \cite{krizhevsky2012imagenet} as the student model for knowledge distillation. For modifying the approach to 3D MTDA, we used PointNet \cite{qi2017pointnet} as a compact student model and PCT \cite{guo2020pct} as a large teacher model. 
`No adaptation' refers to the model trained only by source samples as a na\"ive baseline, and `Supervised' refers to the training performed with labelled target samples.

\paragraph{Evaluation Metric.}
We compare the MTDA performance of the proposed method to the previous works and summarize them in Table~\ref{tab:result1}. 
We use the same pre-processing steps for all methods.
In all the experiments, we report the top-$1$ classification accuracy on the test set, averaged over $3$-folds, for each target domain. 

\subsection{Results and Discussion}
\label{sec:results}

\begin{table*}[ht!]
    \centering
    \caption{Quantitative classification results (\%) on PointDA-10 dataset in MTDA setting. For every source domain, we report performance for each target domain. \textbf{bold} and second best in \underline{underline}. `No adaptation' refers to the model trained only by source samples and `Supervised' denotes the model when trained with labelled target data}.
    \label{tab:result1}
    
    \begin{tabular}{llllllll}
    \toprule
         Source Domain & \multicolumn{2}{l}{ModelNet (M)} & \multicolumn{2}{l}{ScanNet (S*)} & \multicolumn{2}{l}{ShapeNet (S)} & \\
         Src $\to$ Tgt & M $\to$ S* & M $\to$ S & S* $\to$ M & S* $\to$ S & S$\to$M & S$\to$S* & Average \\
         \midrule
        No adaptation (Baseline) & 35.07 & 11.75 & 52.61 & 29.45 & 33.65 & 11.05 & 28.93 \\
        \cdashlinelr{1-8}
        MMD \cite{long2013transfer} & 57.16 & 22.68 & 55.40 & 28.24 & 36.77 & 24.88 & 37.52 \\
        DANN \cite{ganin2015unsupervised} & 55.03 & 21.64 & 54.79 & 37.37 & {\textbf{42.54}} & \underline{33.78} & 40.86 \\
        ADDA \cite{tzeng2017adversarial} & 29.39 & 38.46 & 46.89 & 20.79 & 35.33 & 24.94 & 32.63 \\
        MCD \cite{saito2018maximum} & {\textbf{57.56}} & 27.37 & 54.11 & {\underline{41.71}} & {\underline{42.30}} & 22.39 & {\underline{40.94}} \\
        PointDAN \cite{qin2019pointdan} & 30.19 & {\underline{44.26}} & 43.17 & 14.30 & 26.44 & 28.92 & 31.21 \\  
        \cdashlinelr{1-8}
        AMEAN \cite{chen2019blending} & {\underline{55.73}} & 33.53 & 51.50 & 30.89 & 34.73 & 22.21 & 38.10 \\
        MTDA-ITA \cite{gholami2020unsupervised} & 55.23 & 20.96 & {\underline{56.12}} & 33.71 & 32.33 & 25.62 & 37.33 \\
        MT-MTDA \cite{nguyen2021unsupervised} & 45.43 & 25.72 & 28.25 & 19.51 & 24.65 & \textbf{35.27} & 29.81 \\
        \midrule
        
          \textbf{\methodname (Ours)} &  45.31 & {\textbf{61.36}} & {\textbf{56.67}} & {\textbf{46.63}} & 37.02 & {{27.19}} & {\textbf{45.70}} \\
          ~$\hookrightarrow$ w/o mixup &  28.48 & 40.05 & 33.89 & 12.14 & 27.83 & 24.48 & 27.81 \\
          \midrule
          Supervised in each domain & 77.99 & 67.18 & 79.83 & 66.27 & 63.41 & 53.02 & 67.95 \\
         \bottomrule
    \end{tabular}
\end{table*}

\begin{table*}[h] 
    \centering
    \caption{Quantitative classification results (\%) on PointDA-10 dataset in MTDA setting in different mixup scenarios. For every source domain, we report performance for each target domain. Best result in \textbf{bold} and second best in {\underline{underline}}.}
    \label{tab:result3}
    
    \begin{tabular}{llllllll}
    \toprule
         Source Domain & \multicolumn{2}{l}{ModelNet (M)} & \multicolumn{2}{l}{ScanNet (S*)} & \multicolumn{2}{l}{ShapeNet (S)} & \\
         Src $\to$ Tgt & M $\to$ S* & M $\to$S & S* $\to$ M & S*$\to$ S & S$\to $M & S$\to $S* & Average \\
         \midrule
        {\bf \methodname (Ours)} &  45.31 & {\textbf{61.36}} & {\textbf{56.67}} & {\textbf{46.63}} & {\textbf{37.02}} & 27.19 & {\textbf{45.70}} \\
        Mixup Sep &  41.32 & \underline{47.98} & \underline{{56.18}} & \underline{{42.19}} & 28.85 & \underline{{36.69}} & \underline{{42.20}} \\
        \cdashlinelr{1-8}
        Factor-Mixup &  41.31 & 41.49 & 50.77 & 38.82 & 30.77 & 36.81 & 40.00 \\
        Concat-Mixup &  {\underline{49.20}} & 29.57 & 50.47 & 37.5 & {\underline{33.05}} & 25.64 & 37.57  \\
         Inter-Mixup &  {\textbf{50.95}} & 28.65 & 51.71 & 34.38 & 32.21 & {\textbf{40.80}} & 39.78  \\
          \midrule
         {Best of all methods} & 50.95 & 61.36 & 56.67 & 46.63 & 37.02 & 40.80 & 48.91 \\
         \bottomrule
    \end{tabular}
\end{table*}

We summarize comparative results for classification on PointDA-10 in Table \ref{tab:result1}.
The proposed approach outperforms UDA methods, STDA method for point clouds and MTDA approaches designed for 2D vision modified for 3D point clouds. 
Despite the large domain gap rising due to sim-to-real or real-to-sim adaptation on $M\rightarrow S^{*}$ and $S^{*}\rightarrow M$, respectively, the proposed approach significantly improves the overall performance.

MCD and DANN outperform most of the other methods, but performs worse than our approach.
It is partly because they disentangle the domain-shared features from the domain-specific features, thus achieve better domain generalization.
Moreover, we observe that a simple extension of STDA methods to MTDA does not adapt well on multiple target domains. 
For instance, MMD and DANN achieve an average accuracy of around 42 \% in the STDA setup while they barely reach an accuracy of 40 \% in the MTDA setup.
Interestingly, MCD still performs better than most other methods.
Furthermore, UDA methods designed for point clouds also do not perform well when applied to multiple targets, possibly due to catastrophic forgetting during sequential training on multiple target domains.
We discuss the performance of methods in STDA setup in more detail in Table \ref{tab:result2} of the supplementary due to space sake.

The MTDA methods designed for 2D vision tasks, such as AMEAN, MT-MTDA, and MTDA-ITA, do not perform well on 3D data due to their failure in capturing the local and global geometry of the data while aligning the features across domains.
While methods designed for 2D tasks focus on aligning the global image features, local geometry plays a crucial role in achieving good performance for 3D data~\cite{qin2019pointdan}.
This suggests that modality difference can cause a performance drop due to the inherent property differences of each modality, such as brightness or texture in 2D data and geometry, point density, or orientation in 3D data.
By incorporating local and global geometry information, our approach is able to align features across domains while preserving the intrinsic structures of 3D data, leading to better domain adaptation performance.
Furthermore, the node attention module helps in focussing on important regions of the point cloud, which is critical for accurate classification. 
These design choices allow our model to effectively capture the modality-specific properties of 3D data, resulting in superior performance compared to existing MTDA methods.
For MT-MTDA that uses knowledge distillation, a larger teacher model and a compact student model is desired.
However, if the teacher model fails to align local structures to the global structure, it becomes challenging to transfer accurate knowledge to the student model, leading to relatively disappointing results. 

AMEAN and MTDA-ITA perform better than other MTDA baselines.
MTDA-ITA finds a strong link between the shared latent space common to all domains, while simultaneously accounting for the remaining private, domain-specific factors. 
Whereas AMEAN mixes the target domains and creates sub-targets implicitly blended with each other,  resulting in better performance. 
Nonetheless, our approach outperforms AMEAN, as we takes features focus on learning domain-invariant features that are hard to distinguish from their originating domain. This forces the model to improve its classification performance independent of the domain, resulting in better overall performance.

Additionally, in Table \ref{tab:ablation} of the supplementary, we highlight the importance of each module used in the pipeline by conducting an ablation study on each loss term of $\mathcal{L}_{adv}$ in Equ. \ref{eq:l_subs}.
It can be clearly observed that the mixup module significantly improves performance.
Moreover, we show how adversely the class-imbalance in PointDA-10 affects class-wise classification accuracy in Table \cref{tab:class_res} of the supplementary due to space sake. 
Most classes show satisfactory improvements with our proposed approach except for \emph{Bed}, \emph{Bookshelf} and \emph{Sofa}, which highlights the weakness of our model that neglects the scale information, and when different classes share very similar local structures, the model possibly aligns similar structures across these classes (\eg, large columns contained both by \emph{Lamps} and round \emph{Tables}, small legs in \emph{Beds} and \emph{Sofas} or large cuboidal spaces present in \emph{Beds} and \emph{Bookshelves}).

\subsection{Variants of the Mix-up Methods}
To further investigate the effect of equal weight averaging that is proposed in the \methodname, we vary scaling schemes in the averaging of the mixup representations.
Here, we evaluate three different formulations for mixing, and name it as \emph{Factor}-Mixup, \emph{Concat}-Mixup and \emph{Inter}-Mixup.

\paragraph{Factor-Mixup} 
We mix the feature embeddings from multiple domains together and observe the effect of scaling factor in averaging in Equ. \ref{eq:mensa} as:
\begin{equation}\label{4}
    F_m^{factor} = \lambda F_s + \sum_{i=1}^n \frac{1-\lambda}{n} F_{T_i}.
\end{equation}

\paragraph{Concat-Mixup}
Instead of summing the feature embeddings of the domains, we consider concatenation of the mixups with the intuition of learning the proper weights for each mixup embedding for downstream tasks. 
We use a scaling factor $\lambda$ and $\frac{1-\lambda}{n}$ for balancing between source and targets both in feature and label as:
\begin{equation}
    F_m^{concat} = [\lambda F_s, \frac{1-\lambda}{n} F_{T_1},...,\frac{1-\lambda}{n}F_{T_n}],
    \label{eq:mixb_fm}
\end{equation}  
\begin{equation}
    L_m^{concat} = [\lambda, 2 \frac{1-\lambda}{n},.., N\frac{1-\lambda}{n})],
    \label{eq:mixb_lm}
\end{equation}
where $[\cdot,\cdots,\cdot]$ denotes concatenation operation.


\paragraph{Inter-Mixup} 
In addition to aggregating all the domains together in \methodname, we also consider a linear interpolation of the target domains excluding the $F_s$ for both feature and label as: 
\begin{equation}\label{7}
    F_m^T = \lambda F_{T_1} + (1-\lambda) F_{T_2}.
\end{equation}
\begin{equation}\label{8}
    L_m^T = \lambda L_{T_1} + (1-\lambda) L_{T_2}.
\end{equation}
We devised Inter-Mixup, with the intuition that regularizing the target domains alone should help the model to learn a mapping where it is able to learn the target domain-invariant features promoting better MTDA, thus learning a good separation between the source and target domains in the latent space. 


We compare the performance of the variants with \methodname and Sep., and summarize the results in Table \ref{tab:result3}. 
As Scaler-Mixup is a linear interpolation of all the domains together, the mixed feature representation obtained by Scaler-Mixup has large values in each dimension, which may lead to gradients with large magnitude.
It may hurt the accuracy.
Unlike \methodname and other mixup variants, Concat-Mixup concatenates the feature embeddings from multiple domains.
As the number of domains increases, the shared latent space between the domains mixed becomes smaller.
Therefore, it becomes difficult for the model to learn domain-invariant features across all domains, leading to poor performance among all other variants of mixup.
Interestingly, we observe that mixing the target domains together with the source domain in Inter-Mixup performs better on ScanNet which has real-world samples.
We believe it is because the model is able to learn better domain-invariant features between the real and synthetic domains, as the samples from ScanNet are more sparse and occluded as compared to other domains.
Moreover, we show the visualization of the feature embeddings using t-SNE plots in the supplementary.


\section{Conclusion}

We model the multi target domains as a random variable and propose to mix latent space embeddings of all domains in an ensemble average to encode domain invariant information for the 3D point cloud for the first time in literature. 
The mixed representation helps the domain classifier to learn better domain-invariant features and improve the domain adaptation performance in multi-target domain adaptation set-up. 
We demonstrated the efficacy of our approach on the point cloud DA benchmark dataset of PointDA-10 by showing that our approach significantly outperforms UDA, STDA and MTDA methods proposed for 2D data.

{\small
\bibliographystyle{ieee_fullname}
\bibliography{cvpr}
}

\clearpage
\appendix

\section*{Supplementary Material: \emph{\methodname}: Mix-up Ensemble Average for Unsupervised Multi Target Domain Adaptation on 3D Point Clouds}

\section{PointDA-10 Dataset}
\label{sec:dataset}

\begin{figure}[h!]
    \centering
    \includegraphics[width=0.5\linewidth]{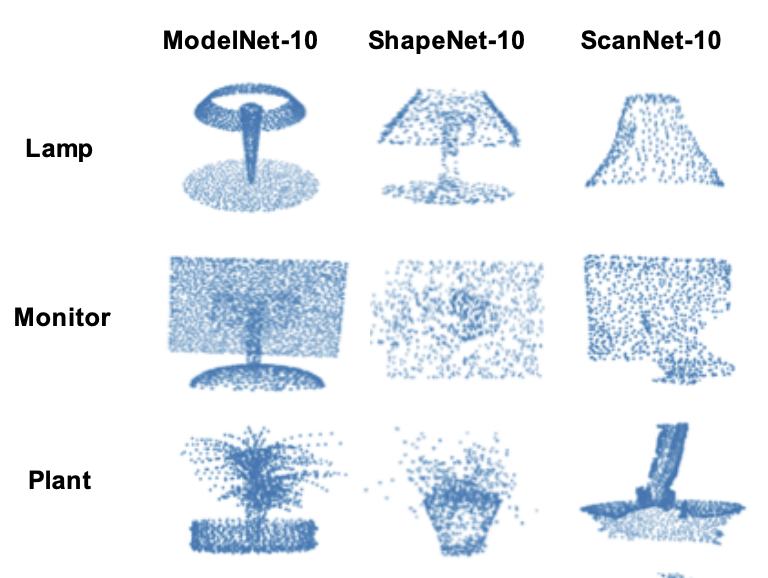}
    \caption{Samples from the PointDA-10 dataset}
    \label{fig:sample}
\end{figure}

The PointDA-10 dataset was proposed for cross-domain 3D objects classification on point clouds~\cite{qin2019pointdan}.
It has been used as a general benchmark for single target domain adaptation (STDA) in literature~\cite{huang2021generation, achituve2021self, qin2019pointdan}.
It consists of subsets of three widely used point cloud datasets: ShapeNet \cite{chang2015shapenet}, ScanNet \cite{dai2017scannet} and ModelNet \cite{wu20153d}.
All three subsets, \ie, ModelNet-10 (M), ScanNet-10 (S*) and ShapeNet (S), share ten common categories (\eg, chair, table and monitor) across them.
We present sample point clouds and statistics of the dataset in Fig \ref{fig:sample} and Table \ref{tab:data}, respectively. 

The dataset is highly class-imbalanced; ModelNet-10 has around 124 Lamp samples while ScanNet-10 and ShapeNet-10 have 161 and 1,620 Lamp samples respectively in the training set.
This causes additional difficulty to adapt to the target domains. 

\begin{table*}[h!]
    \centering
    \caption{Number of samples in PointDA-10 dataset}
    \label{tab:data}
        \begin{tabular}{lllllllllllll}
        \toprule
        \multicolumn{2}{c}{Dataset} &  Bathtub & Bed &  Bookshelf & Cabinet & Chair & Lamp & Monitor & Plant & Sofa  & Table & Total \\
        \midrule
        \multirow{2}{*}{\textbf{M}} & Train  & 106 & 515 & 572 & 200 & 889 & 124 & 465 & 240 & 680 & 392 & 4183 \\
         & Test & 50 & 100 & 100 & 86 & 100 & 20 & 100 & 100 & 100 & 100 & 856 \\
        \midrule
        \multirow{2}{*}{\textbf{S}} & Train & 599 & 167 & 310 & 1076 & 4612 & 1620 & 762 & 158 & 2198 & 5876 & 17378 \\
        & Test & 85 & 23 & 50 & 126 &  662 &  232 & 112 & 30 & 330 & 842 & 2492 \\
        \midrule
        \multirow{2}{*}{\textbf{S*}} & Train & 98 & 329 & 464 & 650 & 2578 & 161 & 210 & 88 & 495 & 1037 & 6110 \\
        & Test & 26 & 85 & 146 & 149 & 801 & 41 & 61 & 25 & 134 & 301 & 1769 \\
        \bottomrule
        \end{tabular}
\end{table*}

\begin{table*}[hbt!]
    \centering
    \caption{Quantitative classification accuracy (\%) on PointDA-10 dataset in STDA set-up reproduced by us and reported from literature \cite{huang2021generation}. No adaptation refers to the model trained only by source samples. The results of the respective methods in the MTDA setup are also reported alongside}.
    \label{tab:result2}
   \begin{tabular}{llllllll} 
    \toprule
        Src $\to$ Tgt & M $\to$ S & M $\to$S* & S* $\to$ M & S*  $\to$ S & S $\to$ M & S $\to$ S* & Average \\
         \midrule
        No adaptation~\cite{huang2021generation} & 42.50 & 22.30 &  39.90 & 23.50 & 34.20 & 46.90 & 34.93 \\
        $\hookrightarrow$ {\footnotesize Reproduced} & 45.52 & 30.79 & 54.90 & 31.37 & 37.26 & 44.50 & 40.72 \\
        $\hookrightarrow$ {\footnotesize in MTDA setup} & 35.07 & 11.75 & 52.61 & 29.45 & 33.65 & 11.05 & 28.93 \\
        \cdashlinelr{1-8} 
        MMD~\cite{long2013transfer} & 57.50 &  27.90 &  40.70 &  26.70 &  47.30 &  54.80 &  42.50 \\
        $\hookrightarrow$ {\footnotesize Reproduced}   & 59.34 & 55.70 & {58.37} & 53.49 & 47.92 & 44.88 & 53.28 \\
        $\hookrightarrow$ {\footnotesize in MTDA setup}  & 57.16 & 22.68 & 55.40 & 28.24 & 36.77 & 24.88 & 37.52 \\
        \cdashlinelr{1-8}
        DANN ~\cite{ganin2015unsupervised} & 58.70 &  29.40 &  {42.30} &  30.50 &  48.10 &  56.70 &  44.20  \\
        $\hookrightarrow$ {\footnotesize Reproduced}  & 50.65 &	54.27 &	54.19 &	52.00 &	48.11 &	47.53 &	51.13 \\
        $\hookrightarrow$ {\footnotesize in MTDA setup}  & 55.03 & 21.64 & 54.79 & 37.37 & {42.54} & {33.78} & 40.86 \\
        \cdashlinelr{1-8}
        ADDA ~\cite{tzeng2017adversarial} & 61.00 &  30.50 &  40.40 &  29.30 &  {48.90} &  51.10 &  43.50 \\
        $\hookrightarrow$ {\footnotesize Reproduced}  & 35.64 & 33.90 & 40.93 & 39.86 & 27.15 &	32.49 &	34.88 \\
        $\hookrightarrow$ {\footnotesize in MTDA setup} & 29.39 & 38.46 & 46.89 & 20.79 & 35.33 & 24.94 & 32.63 \\
        \cdashlinelr{1-8}
        MCD ~\cite{saito2018maximum} & {62.00} & {31.00} & 41.40 & {31.30} & 46.80 & {{59.30}} & {45.30} \\
        $\hookrightarrow$ {\footnotesize Reproduced}  & {62.27} &	{61.21} & {54.25} & 57.59 &	49.76 &	53.46 &	{56.42} \\
        $\hookrightarrow$ {\footnotesize in MTDA setup} & {{57.56}} & 27.37 & 54.11 & {{41.71}} & {{42.30}} & 22.39 & {{40.94}} \\
        \cdashlinelr{1-8}
        PointDAN ~\cite{qin2019pointdan} & {{62.50}}&  {{31.20}} &  {41.50} &  {{31.50}} &  46.90 &  {59.30} &  {{45.50}} \\
        $\hookrightarrow$ {\footnotesize Reproduced}  & 57.57 & 30.63 &	51.80 &	{58.10} &	{51.68} &	25.06 &	45.81\\
        $\hookrightarrow$ {\footnotesize in MTDA setup}  & 30.19 & {{44.26}} & 43.17 & 14.30 & 26.44 & 28.92 & 31.21 \\  
         \bottomrule
    \end{tabular}
    \hfill
\end{table*}

\begin{table*}[hbt!]
    \caption{{\textbf{Ablation}. Quantitative classification accuracy (\%) on the contribution of each module in $\mathcal{L}_{adv}$ as per Eq. 5 (in main paper) towards the overall pipeline (Fig. 2 in main paper) in MTDA setting. Best results are in \textbf{bold} and second best in \underline{underline}}}
    \centering
    \begin{tabular}{llllllll}
    \toprule
        Source Domain & \multicolumn{2}{l}{ModelNet (M)} & \multicolumn{2}{l}{ScanNet (S*)} & \multicolumn{2}{l}{ShapeNet (S)} & \\
         Loss Terms (Eq. 5) & M $\to$ S* & M $\to$S & S* $\to$ M & S* $\to$ S & S$\to $M & S$\to $S* & Average \\
         \midrule
        $\mathcal{L}_{dc}$ & 34.42 & 45.08 & 32.81 & 13.32 & 23.55 & \textbf{38.13} & 31.22 \\ 
        $\mathcal{L}_{mmd}$ & 43.37 & 36.05 & 51.87 & 29.20 & \underline{30.67} & 25.75 & 36.15 \\ 
        $\mathcal{L}_{mix}$ & 32.67 & 43.51 & \textbf{57.88} & \underline{33.17} & 30.52 & 31.59 & 38.22 \\ 
        $\mathcal{L}_{dc}$ + $\mathcal{L}_{mmd}$ & 41.05 & 41.78 & 42.67 & 19.83 & 29.08 & \underline{33.62} & 34.67 \\ 
        $\mathcal{L}_{dc}$ + $\mathcal{L}_{mix}$ & 35.07 & 45.19 & 35.29 & 16.34 & 22.59 & 26.79 & 30.21 \\ 
        $\mathcal{L}_{mmd}$ + $\mathcal{L}_{mix}$ & \underline{43.47} & \underline{53.17} & 55.95 & 30.04 & 28.60 & 30.40 & \underline{40.27} \\ 
        \midrule
        $\mathcal{L}_{dc}$ + $\mathcal{L}_{mix}$ + $\mathcal{L}_{mmd}$ & \textbf{45.31} & \textbf{61.36} & \underline{56.67} & \textbf{46.63} & \textbf{37.02} & 27.19 & \textbf{45.70} \\ 
        \bottomrule
    \end{tabular}
    \label{tab:ablation}
\end{table*}

\begin{table*}
    \centering
    \caption{{Class-wise classification accuracy (\%) on ModelNet to ScanNet in MTDA setting. `No adaptation' refers to the model trained only on Source samples and `Supervised' denotes the model trained with labelled target data.}}
    \label{tab:class_res}
    
    \small
    \begin{tabular}{llllllllllll}
    
    \toprule
        Method & Bathtub & Bed & Bookshelf & Cabinet & Chair & Lamp & Monitor & Plant & Sofa & Table  & Average \\ 
    \midrule
        No adaptation (Baseline) & 40.49 & 21.95 & 12.58 & 6.80 & 11.11 & 46.58 & 51.86 & 56.00 & 65.74 & 46.46 & 35.96 \\ 
        \cdashlinelr{1-12}
        MMD & 55.75 & 9.75 & 18.81 & 0.68 & 37.54 & 30.76 & 46.94 & 52.00 & 77.87 & 75.82 & 40.59 \\ 
        ADDA & 58.71 & 15.40 & 23.28 & 2.68 & 32.87 & 50.07 & 32.95 & 48.00 & 61.53 & 56.6 & 38.21 \\ 
        DANN & 60.42 & 15.85 & 24.47 & 2.72 & 24.77 & 12.82 & 52.03 & 68.00 & 65.75 & 78.42 & 40.53 \\ 
        MCD & 58.72 & 10.97 & 27.97 & 0.68 & 30.01 & 12.82 & 60.33 & 56.00 & 82.59 & 66.06 & 40.62 \\ 
        \cdashlinelr{1-12}
        AMEAN & 58.40 & 19.05 & 17.12 & 7.52 & 45.17 & 36.58 & 54.75 & 40.00 & 84.61 & 72.30 & 43.55 \\ 
        MTDA-ITA & 67.90 & 11.90 & 4.11 & 20.19 & 21.8 & 12.19 & 56.39 & 45.00 & 85.38 & 83.25 & 40.81 \\ 
        MT-MTDA & 59.23 & 5.88 & 24.66 & 4.69 & 32.08 & 14.63 & 66.55 & 48.00 & 78.21 & 72.66 & 40.66 \\ 
        \cdashlinelr{1-12}
        \textbf{MEnsA (Ours)} & 67.11 & 6.58 & 6.77 & 44.89 & 74.09 & 46.05 & 87.92 & 64.55 & 50.00 & 74.47 & 52.24 \\ 
        \midrule
        Supervised in each domain & 91.10 & 69.51 & 61.05 & 89.23 & 99.67 & 80.76 & 91.57 & 51.37 & 94.08 & 81.97 & 81.03 \\ \bottomrule
    \end{tabular}
\end{table*}

\section{Results in STDA set-up}

In Table 1 of the main paper, we compare our method to the methods that are proposed both for STDA setup and for MTDA setup.
For the methods for the STDA setup (\eg, DANN, MCD, MMD, ADDA and PointDAN \cite{long2013transfer, saito2018maximum, ganin2015unsupervised, qin2019pointdan, tzeng2017adversarial}), we implement them either from the scratch when there is no publicly available code repositories (\eg, DANN, MCD and MMD) or modify the authors' code if there is any (\eg, ADDA and PointDAN).
For the methods proposed in the MTDA setup (\eg, MT-MTDA and AMEAN), we use the authors' implementation.

Here, we validate our implementation of the STDA methods by reproducing the results in their original STDA setup.
Specifically, we compare the accuracy of our own implementation to the reported accuracy in the literature in Table \ref{tab:result2} (please compare the first and the second row in each block). 
We observe that our own implementations successfully reproduce the results of the STDA methods in the STDA set-up; in many methods (\eg, MMD, DANN and MCD), our implementation improves the accuracy by a noticeable margin (+10.78\% in average performance of MMD \cite{long2013transfer}, +11.12\% for MCD \cite{saito2018maximum}, and +6.93\% for DANN \cite{ganin2015unsupervised}). 
We attribute the improvements on MMD to the choice of kernels and variance values.
For the MCD, we use additional data augmentations such as jittering, orientation and \etc and they improve the performance by better maximizing the discrepancy between the source and target domains.
For the DANN, we attribute the improvement to better selection of scalar multiplier used in reversing the gradients since the aforementioned details were not explicitly mentioned in the paper~\cite{huang2021generation}.
Our implementation of PointDAN exhibits comparable performance due to lack of rigorous fine-tuning that the authors might have had employed for selecting the weights of the loss components.

However, we observe a significant decrease in average accuracy for ADDA \cite{tzeng2017adversarial}. 
In  S* $\to$ S, our implementation exhibits +10.56\% and in M $\to$ S* and S* $\to$ M, it exhibits small gains.
But there is a significant drop for M $\to$ S, S $\to$ M and S $\to$ S*.
Note that the ADDA method uses pre-training on source domain data to obtain an initialization for learning a domain adapting classifier.
Although the size and the type of the source data for the pre-training affect the domain adaptation performance, it is not well described in the paper~\cite{huang2021generation}.
We used ModelNet-10 for pre-training to reproduce the results of ADDA on PointDA-10.
However, the performance drop, we believe, is due to the pre-training phase.

We also show the accuracy of our implementation in MTDA setup for easy side-by-side comparison to the STDA setup (please compare the second row to third row in each block). 
Note that the results in the third rows are the ones we have reported in Table 1 in the main paper.
The drop in performance of the methods on moving from STDA to MTDA setup highlights the difficulty of adapting on multiple unlabelled targets using a single source domain.
We believe that the difficulty comes from the fact that the model has to adapt to the different target domains in a single phase~\cite{nguyen2021unsupervised}, hence, due to the limited capacity, if it performs well on one target but it does not perform well on the other targets.
The performance gain in overall accuracy while reproducing the aforementioned methods partly validates our implementation for MTDA setup to be credible.

\begin{figure}[!h]
    \centering
    \vspace{-4.5em}
    \includegraphics[width=0.7\linewidth]{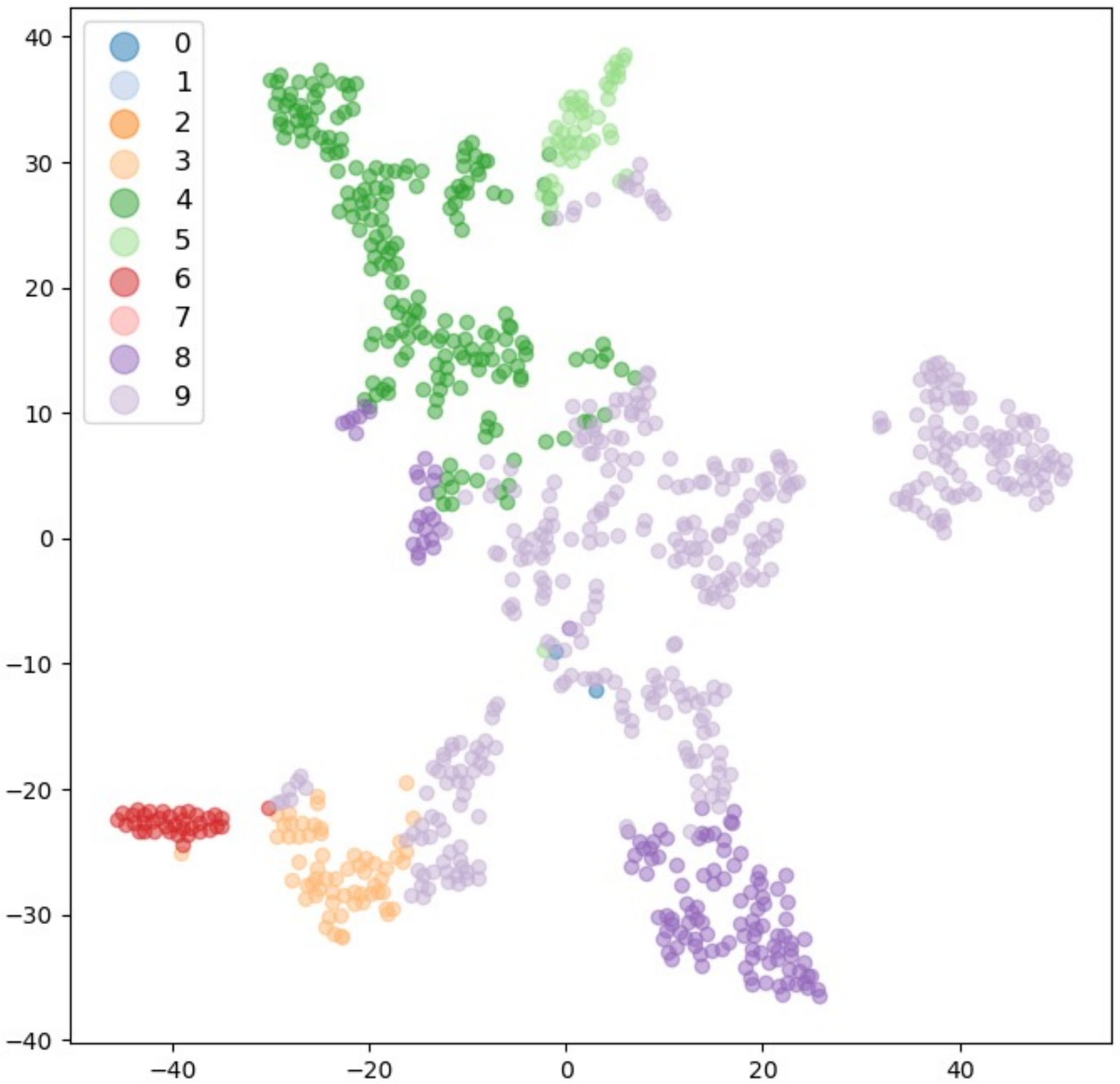}
    \vspace{-3em}
    \caption{t-SNE embedding with perplexity 25 of the proposed method MEnsA on adapting from ShapeNet to ModelNet. Some classes which have distinct shapes are well clustered together. However, some classes with similar geometric structures such as Lamps and Tables, Beds and Sofas, \emph{etc.,} are closer in the cluster.}
    \label{fig:sm_emb}
\end{figure}
\begin{figure}[!ht]
    \centering
    \vspace{-4.5em}
    \includegraphics[width=0.7\linewidth]{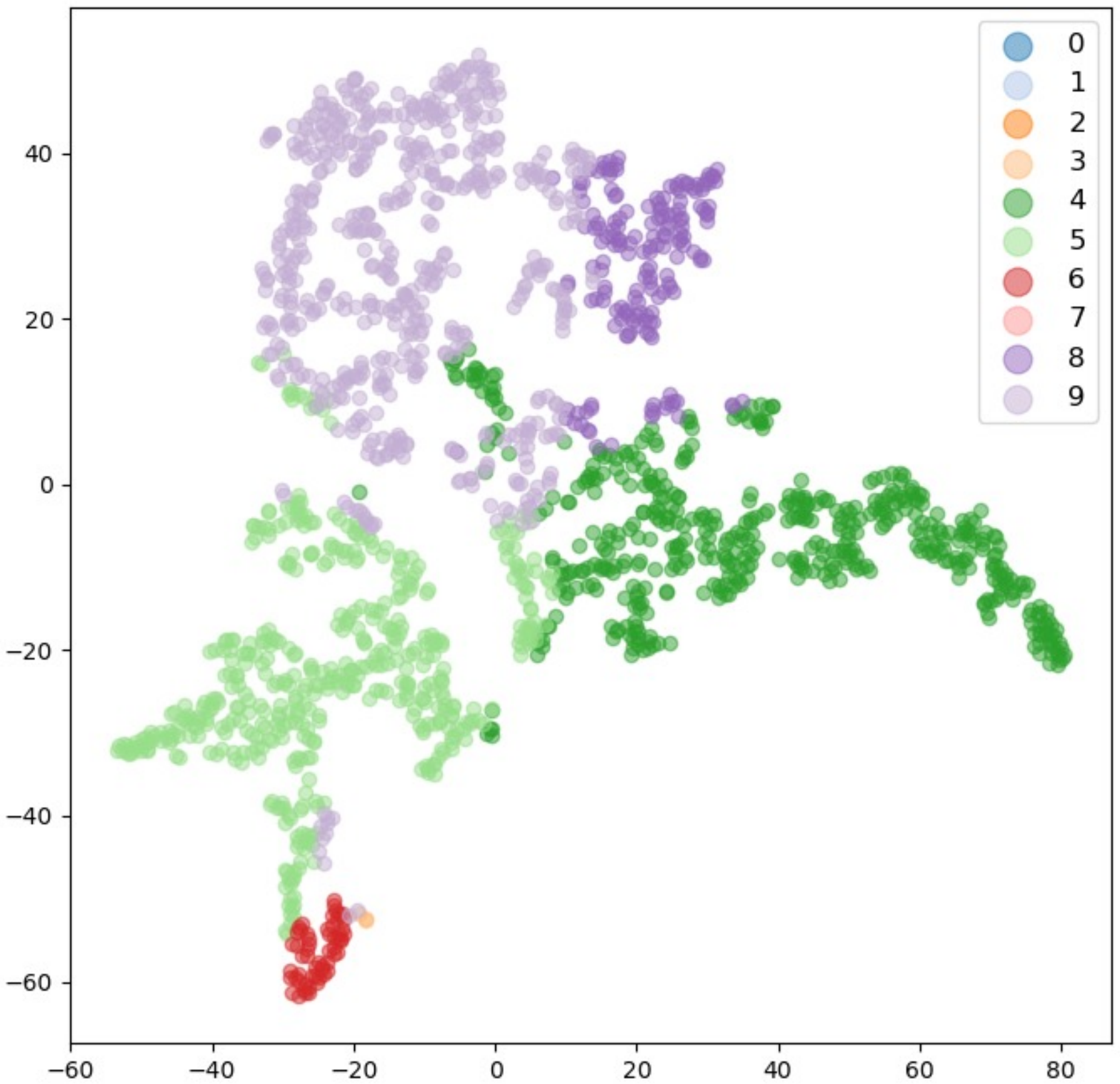}
    \vspace{-3em}
    \caption{t-SNE embedding with perplexity 25 of the proposed method MEnsA on adapting from ShapeNet to Scannet. Here, sim-to-real adaptation is challenging, and the cluster boundaries are not distinct for objects with similar geometric properties.}
    \label{fig:ss_emb}
\end{figure}


\section{Ablation Study}

In Table \ref{tab:ablation}, we ablate $\mathcal{L}_{adv}$ in Eq. 5, which is a linear combination of a domain confusion loss ($\mathcal{L}_{dc}$), an MMD based discrepancy loss ($\mathcal{L}_{mmd}$) and a mixup loss ($\mathcal{L}_{mix}$), for detailed analysis for the contribution of each module, \ie, GRL, MMD, and Domain Mixup module, toward increase in overall accuracy of the proposed method (Fig. 2 of the main paper) in comparison to prior works.
Adversarial domain confusion is implemented using the Gradient Reversal Layer (GRL) ~\cite{ganin2015unsupervised}. The contribution of GRL is measured by $\mathcal{L}_{dc}$.
The GRL helps the model build feature representation of the raw input $\mathcal{X}$ that is good to predict the correct object label $\mathcal{Y}$ subject to the domain label of $\mathcal{X}$ to be not easily deduced by it.
This promotes domain confusion where the feature extractor (\ie, generator) tries to confuse the domain classifier (\ie, discriminator) by bridging the two distributions closer. 
The mapping between the source and target domains is learned via MMD loss, \ie, $\mathcal{L}_{mmd}$.
$\mathcal{L}_{mix}$ controls the flow of information from the proposed domain mixup module.

It is clearly observed from Table \ref{tab:ablation} that the mixup module helps in improving the average classification accuracy as well as accuracy over each domain.
Please note that the proposed approach performs the best when all the three modules are combined coherently as per Eq. 5.

In addition, we conduct a detailed class-wise accuracy analysis across three domains in the PointDA-10 dataset in Table \ref{tab:class_res}. 
We observe decent gains by our method in most of classes over the prior works but not significant gains for \emph{Bed}, \emph{Bookshelf} and \emph{Sofa} classes. 
We believe that the low performance on these classes is due to the fact that the model may neglect the `scale' information; when different classes share very similar local structures, the model possibly aligns similar structures across these classes (\eg, large columns contained both by \emph{Lamps} and round \emph{Tables}, small legs in \emph{Beds} and \emph{Sofas} or large cuboidal spaces present in \emph{Beds} and \emph{Bookshelves}) and leads to classification confusion.







\end{document}